\documentclass[11pt,a4paper]{article}
\usepackage[]{acl2019}
\usepackage{tabularx}
\usepackage{bm}
\usepackage{booktabs}
\usepackage{subcaption}
\usepackage{gb4e}
\usepackage{natbib}

\title{Scalable Cross Lingual Pivots to Model Pronoun Gender for Translation}
\author{Kellie Webster \and Emily Pitler \\
Google Research \\
{\tt \{websterk|epitler\}@google.com}}

\noautomath

\aclfinalcopy

\begin{document}

\maketitle

\begin{abstract}
Machine translation systems with inadequate document understanding can make errors when translating dropped or neutral pronouns into languages with gendered pronouns (e.g., English).
Predicting the underlying gender of these pronouns is difficult since it is not marked textually and must instead be inferred from coreferent mentions in the  context.
We propose a novel cross-lingual pivoting technique for automatically producing high-quality gender labels, and show that this data can be used to fine-tune a BERT classifier with 92\% F1 for Spanish dropped feminine pronouns, compared with 30-51\% for neural machine translation models and 54-71\% for a non-fine-tuned BERT model.
We augment a neural machine translation model with labels from our classifier to improve pronoun translation, while still having parallelizable translation models that translate a sentence at a time.
\end{abstract}

\section{Introduction}
While neural modeling has solved many challenges in machine translation, a number remain, notably document-level consistency.
Since standard architectures are sentence-based, they can misrepresent pronoun gender since cues for this are often extra-sentential \cite{laubli2018has}.
For example, in Figure~\ref{fig:motivating_example},\footnote{Source: \url{https://es.wikipedia.org/wiki/Britney_Spears}, Translation: Retrieved from \url{translate.google.com}, 27 Feb 2019.} there is a dropped subject and a neutral possessive pronoun in the Spanish, which are both translated with masculine pronouns in English despite context indicating they refer to \emph{Britney Jean Spears}.
Recommendations for the use of discourse in translation are developed in \citet{simsmith:2017:DiscoMT}.

Such errors are worrying for their prevalence.
An oracle study found that augmenting an NMT system with perfect pronoun information could give up to 6 BLEU points improvement \cite{stojanovski2018coreference}.
Further, the DiscoMT shared task on pronoun prediction tested multiple language pairs and found that the most difficult language pair by far was Spanish$\rightarrow$English and that ``[t]his result reflects the
difficulty of translating null subjects'' \cite{loaiciga2017findings}. 

\begin{figure}
{\bf Spanish}   \underline{Britney Jean Spears}... {$\bm{\emptyset}$} Adquiri\'o fama durante {\bf su} ni\~nez al participar en el programa de televisi\'on The Mickey Mouse Club (1992). \\
{\bf Spanish$\bm{\rightarrow}$English} {\bf He} gained fame during {\bf his} childhood by participating in the television program The Mickey Mouse Club (1992). 
\\
    \caption{Typical example of pronoun gender errors when translating from pronouns without textual gender marking (e.g., dropped or possessive pronouns in Spanish) to a language with gendered pronouns.}
    \label{fig:motivating_example}
\end{figure}

We address document-level modeling for dropped and neutral personal pronouns in Spanish.
We significantly improve the ability of a strong sentence-level translator to produce correctly gendered pronouns in three stages.
First, we automatically collect high-quality pronoun gender-tagged examples at scale using a novel cross-lingual pivoting technique (Section \ref{sec:data}).
This technique easily scales to produce large amounts of data without explicit human labeling, while remaining largely language agnostic.
This means our technique may also be easily scaled to produce data for other languages than Spanish; we present results on Spanish as a proof of concept. 

Next, we show how to fine-tune BERT \cite{bert2018}, a state-of-the-art pre-trained language model, to classify pronoun gender using our new dataset (Section \ref{sec:bert}).
Finally, we add context tags from our BERT model into standard MT sentence input, yielding a 8.8\% F1 improvement in feminine pronoun generation (Section \ref{sec:mt}).

\section{Background}
In this section, we overview particular nuances associated with pronoun understanding, resulting challenges for translation, and 
existing relevant datasets.

\begin{table}[t]
    \centering
    \begin{tabular}{l|cc}
    \toprule
    Properties  & 
    \multicolumn{2}{c}{Surface Form} \\
              & En & Es \\
              \hline \hline
    Nominative, Masc. & \emph{he} & \emph{\'el} \ \  or $\emptyset$ \\
    Nominative, Fem. & \emph{she} & \emph{ella} or $\emptyset$ \\
    \hline
    Possessive, Masc. & \emph{his} & \emph{su} \\
    Possessive, Fem. & \emph{her} & \emph{su} \\
    \bottomrule
    \end{tabular}
    \caption{Masculine and feminine pronouns cannot be distinguished by their surface forms in Spanish in both the dropped and possessive case.}
    \label{tab:pronountable}
\end{table}

\paragraph{Dropped (Implicit) Pronouns}
Spanish personal pronouns have less textual marking of gender than their English counterparts.
Where they may be inferred from context, Spanish drops subject pronouns \emph{\'el}/\emph{ella} (\emph{he}/\emph{she}).  
Several other languages such as Chinese, Japanese,
Greek, and Italian also have dropped pronouns.

\paragraph{Possessive Pronouns}
Additionally, Spanish possessive pronouns do not mark the gender of the possessor, with \emph{su}(\emph{s}) used where English uses \emph{his} and \emph{her}.
In other Romance languages such as French, the possessive pronoun agrees with the gender of the possessed object,
rather than the possessor as in English.
These usages are summarized for Spanish in Table~\ref{tab:pronountable} and illustrated in Figure~\ref{fig:motivating_example}.

\paragraph{MT of Ambiguous Pronouns}
Document-level consistency is an outstanding challenge for machine translation \cite{laubli2018has}.
One reason is that standard techniques translate the sentences independently of one another, meaning that context outside a given sentence cannot be used to make translation decisions.
In a language like Spanish, this means that critical cues for pronoun gender are simply not present at decoding time.

There have been a number of directions exploring how to introduce general contextual awareness into translation.
\citet{tiedemann-scherrer:2017:DiscoMT} investigated a simple technique of providing multiple sentences, concatenated together, to a translator and learning a model to translate the final sentence in the stream using the preceding one(s) for contextualization.
Unfortunately, no substantial performance gain was seen.
Next, the Transformer model \cite{vaswani2017} was extended by introducing a second, context decoder \cite{wang-EtAl:2017:EMNLP20179,zhang-EtAl:2018:EMNLP2}.
Complementary to this, \citet{kuang-EtAl:2018:C18-1} introduces two caches, one lexical and the other topical, to store information from context pertinent to translating future sentences.

Specific to the translation of implicit or dropped pronouns, \citet{wang-EtAl:2018:EMNLP11} shows that modeling pronoun prediction jointly with translation improves quality, while \citet{miculicichwerlen-popescubelis:2017:CORBON} presents a proof-of-concept where coreference links are available at decoding time.
Most similar to our current work, \citet{habash-etal-2019-automatic,moryossef-etal-2019-filling,vanmassenhove-etal-2018-getting} improve gender fidelity in translations in of first-person singular pronouns in languages with grammatical gender inflection in this position.
They show promising results from either injecting explicit information about speaker gender, or predicting it with a classifier.
We extend this line of work without requiring human labels to train our classifier.

\begin{table*}[t]
    \centering
    \begin{tabular}{lp{11cm}}
    \toprule
 \multicolumn{2}{l}{\textbf{(A) Page Alignment}}
 \\
    Title: "Mitsuko Shiga" & \url{https://en.wikipedia.org/wiki/Mitsuko_Shiga} \\
                           & \url{https://es.wikipedia.org/wiki/Mitsuko_Shiga} \\
    \hline
    \multicolumn{2}{l}{\textbf{(B) Sentence Alignment}} \\
     Spanish     & Public\'o numerosas antolog\'ias de {\bf \it su} poesía durante {\bf \it su} vida, incluyendo Fuji no Mi , Asa Tsuki, Asa Ginu, y Kamakura Zakki.
      \\
   Spanish$\rightarrow$English & {\bf  He} published numerous anthologies of {\bf  his} poetry during {\bf  his} life, including Fuji no Mi, Asa Tsuki, Asa Ginu, and Kamakura Zakki. \\
     English   &  {\bf  She} published numerous anthologies of {\bf  her} poetry during {\bf  her} lifetime, including Fuji no Mi ("Wisteria Beans"), Asa Tsuki ("Morning Moon"), Asa Ginu ("Linen Silk"), and Kamakura Zakki ("Kamakura Miscellany").\\
    \hline
    \textbf{(C) Pronoun Tagging} &
    {\bf $\bm{\emptyset}$/She} Public\'o numerosas antologías de {\bf su/her} poesía durante {\bf su/her} vida, incluyendo Fuji no Mi , Asa Tsuki, Asa Ginu, y Kamakura Zakki.\\
    \bottomrule
    \end{tabular}
    \caption{Example of how pronoun translations are extracted from comparable texts.  The English and Spanish sentences were extracted from Wikipedia articles about the same subject, the sentences were aligned due to the similarities in their translations, and the labels were extracted by matching the Spanish dropped and ambiguous pronouns with the English pronouns.}
    \label{tab:dataexample}
\end{table*}

\paragraph{Prior Work on Aligned Corpora}
Translation of sentences has been a task at the Workshop on Machine Translation (WMT) since 2006.
Over this period, more and larger datasets have been released for training statistical models.
Typically released data is given in sentence-pair form, meaning that for each source language sentence (e.g. Spanish), there is a target language sentence (e.g. English) which may be considered its true translation.
For a discussion of the adequacy of this approach, see \citet{guillou-hardmeier:2018:EMNLP}.

Translation of Spanish sentences to English was most recently contested at WMT'13\footnote{\url{http://www.statmt.org/wmt13/translation-task.html}} \cite{wmt13}, which offered participants 14,980,513 sentence pairs from Europarl\footnote{\url{http://www.statmt.org/europarl/}} \cite{europarl2017}, Common Crawl \cite{smith-EtAl:2013:ACL2013}, the United Nations corpus\footnote{\url{https://cms.unov.org/UNCorpus/}} \cite{un2016}, and news commentary.
While large, this dataset wasn't designed for document-level understanding.
Document boundaries are available, but correspond loosely to what a user of a translation system might consider a document, whole days of European parliament, for instance.

DiscoMT was introduced as a workshop in 2013\footnote{\url{http://aclweb.org/anthology/W13-3300}} to target the translation of pronouns in document context.
As well as Europarl, Ted talk subtitles\footnote{\url{http://opus.nlpl.eu/TED2013.php}} \cite{tedtalks2012} were included for both training and evaluation.
Also available for the translation of extended speech is Open Subtitles\footnote{\url{http://opus.nlpl.eu/OpenSubtitles.php}} \cite{opensubtitles2016,mller-EtAl:2018:WMT}, which contains utterance pairs which are aligned and contextualized according to their temporal appearance in a film.
Although subtitle datasets make available very wide context, they are not ideal in this study due to their heavy use of first and second person pronouns and noisy alignment.

High quality targeted datasets for discourse phenomena were explored in \citet{bawden-EtAl:2018:N18-1}, however the released corpus was a small French$\rightarrow$English test set.
In this paper, we introduce a novel cross-lingual technique for gathering a large-scale targeted dataset without explicit human annotation.

\section{Cross-Lingual Pivoting for Pronoun Gender
Data Creation}
\label{sec:data}

Despite document-level context being critical for pronoun translation quality, little of the available aligned data is directly suitable for training models.
In this section, we exploit the special properties of Wikipedia to automate the collection of high-quality gender-tagged pronouns in document context.

\begin{table*}[t]
    \centering
    \begin{tabular}{l|rrr|rrr}
    \toprule
         & \multicolumn{3}{c|}{Prodrop} & \multicolumn{3}{c}{Possessive} \\
         & Articles & \multicolumn{1}{c}{he} & \multicolumn{1}{c|}{she} & Articles & \multicolumn{1}{c}{his} & \multicolumn{1}{c}{her} \\
    \hline
    \hline
    Spanish Wikipedia  & 1,326,469 & - & - & 1,326,469 & - & - \\
    Page Alignment     &   420,850 & - & - &   420,850 & - & - \\
    Pronoun Tagging    &    45,210 & 118,911 & 41,524 & 52,559 & 266,886 & 111,411 \\
    \bottomrule
    \end{tabular}
    \caption{Extraction yield in each stage of the multi-lingual pivot extraction pipeline.}
    \label{tab:yield}
\end{table*}

\subsection{Cross-Lingual Pivoting}

Wikipedia has a number of desirable properties which we exploit to automatically tag pronoun gender.
Firstly, it is very large for head languages like English and Spanish.
More importantly, many non-English Wikipedia pages express similar content to their English counterpart given the resource's style and formatting requirements, and the ease of using translation systems to generate base text for a missing non-English page. 
Based on these properties, we develop the following three stage extraction pipeline, illustrated in Table~\ref{tab:dataexample} with pipeline yield summarized in Table~\ref{tab:yield}.

\paragraph{Page Alignment} Identify as pairs pages with the same title in English and Spanish. We prioritize precision and require exact match.

\paragraph{Sentence Alignment} Identify sentence pairs  in aligned pages which express nearly the same meaning. 
To do this, we compare English translations of sentences\footnote{From an in-house tool.} from the Spanish page to sentences from the English page. 
We do bipartite matching over these English sentence-pairs, selecting pairings with the smallest edit distance. 
We require that edit distance is maximally one half sentence length, that paired sentences share either a noun or verb, and that the mapping is at most one-to-one.
Table \ref{tab:dataexample}, step (B) shows an aligned sentence pair.

\paragraph{Pronoun Tagging} Perform alignment\footnote{We use an implementation of \citet{alignment1996}.} over the tokens in detected sentence pairs, identifying where English uses a gendered pronoun and Spanish has either a dropped or possessive pronoun.
Copy the gender of the English pronoun as a label onto the ambiguous target pronoun, masculine for \emph{he} and \emph{his}, feminine for \emph{she} and \emph{her}.
Step (C) in Table \ref{tab:dataexample} shows the resulting pronoun tagging in the example.

\subsection{Final Dataset}

\paragraph{Gender} We can see from Table~\ref{tab:yield} that there are approximately three times as many masculine examples extracted compared to feminine examples.
This is not surprising given the disparity in representation by gender found in Wikipedia \cite{wagner2016}.
For the final dataset, we down-sample masculine examples to achieve a 1:1 gender balance, yielding a final dataset with 79,240 prodrop and 187,224 possessive examples.
We split the 79,240 prodrop examples into training, development, and test sets of size 72,120, 2,968, and 4,152, respectively;
the 187,224 possessive examples are similarly split into sets of size 167,222, 8,862, and 11,140.

\paragraph{Quality}

To assess the quality of our examples, we sent 993 examples for human rating.
In-house bilingual speakers were shown a paragraph of Spanish text containing a labeled pronoun and asked whether, given the context, the pronoun referred to a masculine person, feminine person, or whether it was unclear.
Agreement between these human labels and those produced by our automatic pipeline is high.
84\% of prodrop and 80\% of possessive pronouns could be disambiguated with the provided context   and, of those, 98\% and 92\% of automatically generated gender tags agreed with human judgment.
Where the provided paragraph was not enough, it was typically the case that relevant context was in the directly preceding paragraph.
Confusion matrices are given in Tables~\ref{tab:prodrop_human} and \ref{tab:possessive_human}.

\begin{table}[]
    \centering
    \begin{tabular}{ll|rrr}
    \toprule
        & & \multicolumn{3}{c}{Human} \\
        & & he & she & unclear \\
    \hline
    \hline
    Pipeline 
    & he  & 221 &   3 & 36 \\
    & she &   7 & 179 & 43 \\
    \bottomrule
    \end{tabular}
    \caption{Agreement between automatic labels and human judgment for prodrop examples.}
    \label{tab:prodrop_human}
\end{table}

\begin{table}[]
    \centering
    \begin{tabular}{ll|rrr}
    \toprule
        & & \multicolumn{3}{c}{Human} \\
        & & his & her & unclear \\
    \hline
    \hline
    Pipeline
    & his & 198 &   6 & 48 \\
    & her &  25 & 179 & 55 \\
    \bottomrule
    \end{tabular}
    \caption{Agreement between automatic labels and human judgment for possessive pronoun examples.}
    \label{tab:possessive_human}
\end{table}

\begin{table*}[]
    \centering
    \begin{tabular}{ll|rrr|rrr}
    \toprule
            &   & \multicolumn{3}{c|}{Masculine}
                & \multicolumn{3}{c}{Feminine} \\
            &   & \multicolumn{1}{c}{P} & \multicolumn{1}{c}{R} & \multicolumn{1}{c|}{F1} & \multicolumn{1}{c}{P} & \multicolumn{1}{c}{R} & \multicolumn{1}{c}{F1} \\
    \hline
    \hline
    Prodrop     
    & Baseline MT (Sentences) & 56.2 & 81.3 & 66.5 & 91.3 & 18.6 & 30.9 \\
    & Baseline MT (Contexts) & 62.4 & 60.5 & 61.4 & 93.0 & 25.1 & 39.5 \\
    & Context MT (Contexts) & 65.1 & 65.3 & 65.2 & 88.9 & 35.8 & 51.0 \\
    & BERT (Sentences) & 66.6 & 61.6 & 64.0 & 87.0 & 40.2 & 54.9 \\
    & BERT (Contexts) & 81.6 & 58.5 & 68.1 & 92.6 & 58.5 & 71.7 \\
    & BERT (Contexts) + Data & \bf 90.4 & \bf 95.2 & \bf 92.7 & \bf 95.0 & \bf 89.8 & \bf 92.3 \\
    \hline
    Possessives 
    & Baseline MT (Sentences) & 58.2 & 77.8 & 66.6 & 86.7 & 30.4 & 45.0 \\
    & Baseline MT (Contexts) & 63.6 & 63.4 & 63.5 & 87.2 & 35.0 & 50.0  \\
    & Context MT (Contexts) & 64.2 & 67.6 & 65.9 & 86.4 & 38.0 & 52.8 \\
    & BERT (Contexts) + Data & \bf 87.4 & \bf 91.3 & \bf 89.3 & \bf 90.9 & \bf 86.8 & \bf 88.8 \\
    \bottomrule
    \end{tabular}
    \caption{Intrinsic evaluation of pronoun gender classification over our new test sets.}
    \label{tab:intrinsic}
\end{table*}

\section{Classifying Pronoun Gender}
\label{sec:bert}

Language model (LM) pretraining \cite{elmo2018,bert2018} is emerging as an essential technique for human-quality language understanding.
However, LMs are limited to learning from what is explicitly expressed in text, which seems insufficient for Spanish given that gender is often not textually realized for pronouns.
We show that BERT, a state-of-the-art pretrained-LM model \cite{bert2018}, already models some aspects of pronoun gender, achieving scores above a strong NMT baseline.
We then show that fine-tuning this LM using our new data improves pronoun gender discrimination by over 20\% absolute F1 for both genders.

\subsection{Neural Machine Translation}

\begin{table}[]
    \centering
    \begin{tabular}{lrr|r}
    \toprule
     & Count & Prodrop & Rate \\
    \hline
    \hline
    he  & 1,220 & 1,106 & 90.7\% \\
    she &   314 &   251 & 79.9\% \\ 
    \bottomrule
    \end{tabular}
    \caption{Even accounting for difference in the raw number of pronouns by gender, prodrop is more likely for masculine entities than feminine entities in WMT'13 (development).}
    \label{tab:prodrop}
\end{table}

We take as our NMT baseline a Transformer \cite{vaswani2017} with vocabulary size 32,000 trained for up to 250k steps.
We set input dimension to 512 and use 6 layers with 8 attention heads and hidden dimension 2048, dropout probabilities of 0.1 and learning rate 2.0.
We train this model on WMT'13 Spanish$\rightarrow$English data,\footnote{\url{http://www.statmt.org/wmt13/translation-task.html}} preprocessed using the Moses toolkit.\footnote{\url{https://github.com/moses-smt/mosesdecoder}}
Rows designated Baseline in Table~\ref{tab:intrinsic} are trained with the standard sentence-level formulation of the task and yield a best test set BLEU score \cite{papineni-EtAl:2002:ACL} of 34.02.
Context MT is trained using the 2+1 strategy of \cite{tiedemann-scherrer:2017:DiscoMT} where an extended context, in our case full sentences up to 128 subtokens,\footnote{Given this very local context, we make the assumption that adjacent sentences in the pre-shuffled dataset are document-continuous. This assumption introduces noise at document boundaries, which occur much less frequently than document continuation transitions.} is used to predict the translation of the contextualized target sentence.
This modeling yields 33.99 BLEU, which is not substantially different to Baseline MT, consistent with \citeauthor{tiedemann-scherrer:2017:DiscoMT}.

To evaluate how well these NMT systems model pronoun gender, we use them to translate the Spanish text from our new dataset into English.
For each target Spanish pronoun, we use an implementation of \citet{alignment1996} to find the corresponding English pronoun and count the translation as correct if the gender tag on the Spanish agrees with the English pronoun gender.
Table~\ref{tab:intrinsic} shows the results when the input units are single sentences (Sentences) or full sentences up to 128 subtokens (Contexts).
All context sentences precede the one containing the pronoun (no lookahead).

Performance is weak overall but especially on feminine examples of prodrop.
The tendency is for masculine pronouns to be over-used:
feminine pronouns are used by Baseline MT (Sentences) less than one time in five when they should be.
To understand why this effect is so strong, we analyzed the development set of WMT'13 (Table~\ref{tab:prodrop}).
Specifically, we counted the number of times the English pronouns \emph{he} and \emph{she} occurred (first column).
Not surprisingly, there was an over-representation of masculine pronouns;
for each feminine example in the data, there was four masculine examples.
Next, we looked in the aligned Spanish sentences for the corresponding pronouns \emph{\'el} and \emph{ella} and counted a case of prodrop each time they were missing (second column).
Even accounting for differences in raw frequency by gender, there is a difference in which prodrop is more frequent for masculine entities.
This is the first time this difference has been reported and future work should control for the prior it induces.

Introducing context in the input to an MT model, either trained on sentences or contexts, improves overall performance by enhancing feminine recall and masculine precision at the expense of masculine recall.
One possible reason is that the added context includes the relevant cues for breaking out of the prior for masculine pronouns.
The magnitude of change in F1 score is perhaps beyond expected given that Baseline~MT and Context~MT achieve similar BLEU scores.
However, this finding builds on the arguments in \citet{cfka-bojar:2018:Long}, which show that BLEU score does not correlate perfectly with semantics. 

\begin{figure}[t]
{\bf Input:} Britney Jean Spears (McComb, Misisipi; 2 de diciembre de 1981), conocida como Britney Spears, es una cantante, bailarina, compositora, modelo, actriz, dise\~nadora de modas y empresaria estadounidense. Comenz\'o a actuar desde ni\~na, a trav\'es de papeles en producciones teatrales. {\bf ***} adquiri\'o fama durante su ni\~nez al participar en el programa de televisi\'on The Mickey Mouse Club (1992). \\
{\bf MaskLM Output:} 
spears (0.61), britney (0.23), tambien (0.06), {\bf ella (0.03)}, rapidamente (0.01)... \\
{\bf Prediction:}  {\bf Feminine} 
\caption{Example of how the BERT pretrained model \cite{bert2018} is evaluated on gendered pronoun prediction.  The MaskLM predicts a word for the wildcard position.  The feminine \emph{ella} is found near the top of the k-best list, so the prediction is \emph{Feminine}.} 
\label{fig:bertprobe}
\end{figure}

\subsection{Language Model Pretraining}

We initialize our LM experiments from a BERT$_{BASE}$ model (110M parameters; see \citet{bert2018}) trained over Wikipedia from 102 languages including Spanish.
To assess how much this model knows about pronoun gender out-of-the-box, we extend the masked LM approach from the original BERT work, inserting a mask token in each dropped subject position (see Figure \ref{fig:bertprobe} for an example).\footnote{This is applied to prodrop only. No analogous probing may be done for the neutral Spanish possessive pronouns without potentially semantic-breaking re-writing.}
To generate gender predictions, we deduce masculine if \emph{\'el} appears before \emph{ella} in the list of mask fills and vice versa;
we make no gender prediction where neither pronoun appears in the top ten mask fills.
The results of this probing is shown in the BERT rows of Table~\ref{tab:intrinsic}, which use input feeds of one sentence (Sentences) and up to 128 subtokens (Contexts).

Despite scanning the full top ten mask fills, we can see that the strength of BERT is in its strong precision rather than boosts to recall.
While not directly comparable to MT values, which are generated single-shot, we can see that BERT values are stronger, particularly for feminine examples which are handled poorly by machine translation.

\begin{figure*}
\begin{exe}
    \ex 
    \gll Comenzo a actuar desde ni\~na, {a traves de} papeles en {producciones teatrales}. \textbf{$<$t$>$ Adquirio $<$/t$>$} fama durante su ni\~nez. \\
    {S/he started} to act {as a} child.FEM, through roles in {theatrical productions}. {S/he acquired} fame in his/her childhood. \\
    \end{exe}
    \label{fig:lime_input}
\end{figure*}

\begin{table*}[]
    \centering
    \begin{tabular}{clr|clr}
    \toprule
    \multicolumn{3}{c|}{Before Fine-tuning} & \multicolumn{3}{c}{After Fine-tuning} \\
    Class & Token & Weight & Class & Token & Weight \\
    \hline
    \hline
    *\textit{Masc.} & durante & 0.01 & \textbf{Fem.} & ni\~na & 0.29 \\
    & traves  & -0.01 & & Comenzo & -0.10 \\
    & actuar  & -0.01 & & ni\~nez   &  0.09 \\
    & Comenzo &  0.01 & & papeles & -0.01 \\  
  \bottomrule
    \end{tabular}
    \caption{LIME analysis for a representative Spanish text in which all third-person singular pronouns are either dropped or genderless. An English gloss, including grammatical gender marking, is given above.}
    \label{tab:lime}
\end{table*}

\subsection{BERT Classification Task}

We extend this understanding of pronoun gender using the fine-tuning recipe given in the original BERT paper.
Specifically, we generate classification examples where the output is a gender tag (masculine or feminine) and the input text is up to 128 subtokens including special \texttt{<t>} tokens to mark the target pronoun position (as in Table~\ref{tab:lime}).
We train over both prodrop and possessives examples from the training portion of our new dataset.

Table~\ref{tab:intrinsic} shows that this approach is extraordinarily powerful for modeling pronoun gender in our dataset.
We note that the performance we see here is remarkably similar to human ratings of quality for the dataset.

On all metrics and in both genders, we see leaps of 20 points and higher.
There is minimal performance difference across the two genders;
precision and recall do not need to trade off against one another.
Our novel method which allows implicit pronoun gender to be made explicit and amenable to text-level modeling is very well suited to pronoun gender classification.

\subsection{Analysis}

We use the LIME toolkit\footnote{\url{https://github.com/marcotcr/lime}} \cite{lime} to understand how the fine-tuning recipe yields such a strong model of pronoun gender.
LIME is built for interpretability, learning linear models which approximate heavy-weight architectures over features which are intuitive to humans.
By analyzing the highly-weighted features in these approximations, we can learn what factors are important in the classification decisions of the heavy-weight models.

Table~\ref{tab:lime} shows the top tokens in the LIME analysis of our gender classifier, both at the start and end of fine-tuning, for a representative Spanish text, shown with its English gloss.
Before fine-tuning, the classification decision is incorrect and importance is low and spread fairly evenly across all terms in context.
After fine-tuning, a correct classification is made, primarily with reference to the \emph{ni\~na} token, which is the feminine form of \emph{child} and refers to a young Britney Spears, the dropped subject of \emph{Adquiri\'o}.
That is, just as the original Transformer model implicitly models coreference for translation \cite{voita-EtAl:2018:Long,webster2018gap}, a fine-tuned BERT classifier appears to implicitly learn coreference to make gender decisions here.

\section{Improving Machine Translation with Classification Tags}
\label{sec:mt}

\begin{table*}[]
    \centering
    \begin{tabular}{l|l|l|r|rrr|rrr}
    \toprule
    Model & Translation & Classifier & &
    \multicolumn{3}{c|}{Masculine} & \multicolumn{3}{c}{Feminine} \\
          & \multicolumn{1}{c|}{Input} & \multicolumn{1}{c|}{Input} 
          & BLEU & \multicolumn{1}{c}{P} & \multicolumn{1}{c}{R} & \multicolumn{1}{c|}{F1} & \multicolumn{1}{c}{P} & \multicolumn{1}{c}{R} & \multicolumn{1}{c}{F1} \\
    \hline
    \hline
    Baseline MT & Sentences & - & 34.02 & 95.2 & \bf 97.1 & 96.2 & 69.7 & 57.5 & 63.0 \\
    \hspace{0.5cm} + Gender Tags & Sentences & Contexts & \bf 34.12 & 96.8 & 96.2 & \bf 96.5 & \bf 70.0 & 73.7 & \bf 71.8 \\
    Context MT  & Contexts  & - & 33.99 & \bf 97.6 & 94.7 & 96.1 & 63.2 & \bf 80.0 & 70.6 \\
    \bottomrule
    \end{tabular}
    \caption{BLEU score and pronoun accuracy by gender on the WMT'13 Spanish-to-English test set.}
    \label{tab:bleu}
\end{table*}

Given the strong intrinsic evaluation, we use the gender classification tags directly in a standard NMT architecture to introduce implicit contextual awareness while still having translation parallelizable across sentences.
This follows precedent set in \citet{johnson-etal-2017-googles} and \citet{zhou-etal-2019-examining} for controlling translations via tag-based annotations.
Our simple approach yields substantial improvements in pronoun translation quality.

\paragraph{Baseline MT + Gender Tags} We use the Transformer model specification above but train with WMT'13 sentence pairs augmented with gender tags.
For each Spanish sentence, we classify whether it contains a third-person singular dropped subject or possessive pronoun using heuristics over morphological analysis, part of speech tags, and syntactic dependency structure \cite{andor-EtAl:2016:P16-1}.\footnote{Dropped pronouns occur where a \texttt{ROOT} verb has no \texttt{nsubj} dependent and no noun to its left; possessive pronouns are tagged with a label ending in \texttt{\$}.}
For those containing a target pronoun, we concatenate the gender tag predicted by BERT~(Contexts)~+~Data, after the sentence,\footnote{To preserve positional encoding.} following a special \texttt{<c>} context token.  For example:

\begin{exe}
\ex Adquiri\'o fama durante su ni\~nez al participar en el programa de televisi\'on The Mickey Mouse Club (1992). $\textless$c$\textgreater$ $\textless$FEM$\textgreater$ 
\end{exe}

For robustness, we introduce noise by randomly flipping whether the sentence contains a target pronoun 2\% of the time\footnote{In which case, input contains no special \texttt{<c>} token.} and generating a random gender tag 5\% of the time.

We choose this strategy as a simple extension of WMT which is fit for our purposes.
Specifically, we seek to show that our gender classification model captures implicit pronoun gender well and that its understanding is useful for the translation of ambiguous pronouns.

\paragraph{Results} We evaluate translation performance using BLEU score and pronoun generation quality using F1 score disaggregated by gender over gendered pronouns.
We do not report APT score\footnote{\url{https://github.com/idiap/APT}} \cite{miculicich2017validation} since we target few pronouns and make conclusions at the level of gender, not pronoun.

Table~\ref{tab:bleu} shows that incorporating contextual awareness, either from concatenating full sentences or gender tags in translation input, improves pronoun quality markedly.
Masculine performance is uniformly strong and our new model, Baseline~MT~+~Gender~Tags, gives the best performance on feminine pronouns, achieving a 8.8\% F1 score improvement compared to Baseline~MT.
As for the contextual MT models analyzed in the previous section, the improvement is from addressing the low recall over feminine pronouns in sentence-level translation (while retaining precision).
We manually reviewed error cases and noted the major class was from domain shift:
where antecedents were close to pronouns in Wikipedia, there were more frequent cases in WMT of antecedents being outside the BERT context window, and these were misclassified.
Future work could look at methods for learning pre-trained language models from extended document context.

\section{Conclusion}

In this paper, we introduce a cross-lingual pivoting technique which generates large volumes of labeled data for pronoun classification without explicit human annotation.
We demonstrate improvements on translating pronouns from Spanish to English 
using a BERT-based classifier fine-tuned on this data, which achieves F1 of 92\% for feminine  pronouns, compared with
30-51\% for neural machine translation models 
and 54-71\% for non-fine-tuned BERT.
We show that our classifier can be incorporated into a standard sentence-level neural machine translation system to yield 8.8\% F1 score improvement in feminine pronoun generation.

Our data creation method is largely language-agnostic and may be extended in future work to new language pairs, or even different grammatical features.
Further in this direction, we are excited by the prospect of zero-shot pronoun gender classification, particularly in low-resource languages.

\section*{Acknowledgements} 
We would like to thank Melvin Johnson, Romina Stella, and Macduff Hughes for assistance with machine translation work, as well as Mike Collins, Slav Petrov, and other members of the Google Research team for their feedback and suggestions on earlier versions of this manuscript.

\bibliographystyle{acl_natbib}
\bibliography{main}

\end{document}